%% file: arxiv.tex
\documentclass[]{databricksai}
\input{header}
\usepackage{times}
\usepackage{url}            %
\usepackage{booktabs}       %
\usepackage{amsfonts}       %
\usepackage{nicefrac}       %
\usepackage[most]{tcolorbox} % Loads the tcolorbox package
\usepackage{microtype}      %
\usepackage{mkolar_definitions}
\usepackage{xspace}
\usepackage{multirow}
\usepackage{amsmath}
\usepackage{graphicx}
\usepackage{subcaption}
\usepackage{algorithm}
\usepackage{tikz}
\usepackage{color, colortbl}
\usepackage{enumitem}
\usepackage{comment}
\usepackage{bm}
\usepackage{algorithm}
\usepackage{algpseudocode}

\usepackage[scaled=.9]{helvet}

\usepackage{url}

\usepackage{csquotes}

\usepackage[dvipsnames]{xcolor}

\usepackage{natbib}
\bibpunct{(}{)}{;}{a}{,}{,}

\usepackage{centernot}

\newcount\Comments  %
\definecolor{darkgreen}{rgb}{0,0.5,0}
\definecolor{darkred}{rgb}{0.7,0,0}
\definecolor{teal}{rgb}{0.3,0.8,0.8}
\definecolor{orange}{rgb}{1.0,0.5,0.0}
\definecolor{purple}{rgb}{0.8,0.0,0.8}
\definecolor{abstract}{RGB}{208,154,165}
\newcommand{\kibitz}[2]{\ifnum\Comments=1{\textcolor{#1}{\textsf{\footnotesize #2}}}\fi}

\definecolor{Gray}{gray}{0.9}

\input{notation.tex}

\title{LLMs Can Learn to Reason Via Off-Policy RL}
\author[1]{Daniel Ritter}
\author[1,2]{Owen Oertell}
\author[1]{Bradley Guo}
\author[2]{Jonathan D. Chang}
\author[3]{Kianté Brantley}
\author[2]{Wen Sun}

\affiliation[1]{Cornell University}
\affiliation[2]{Databricks}
\affiliation[3]{Harvard University}
\abstract{
Reinforcement learning (RL) approaches for Large Language Models (LLMs) frequently use on-policy algorithms, such as PPO or GRPO. However, policy lag from distributed training architectures and differences between the training and inference policies break this assumption, making the data off-policy by design. To rectify this, prior work has focused on making this off-policy data appear more on-policy, either via importance sampling (IS), or by more closely aligning the training and inference policies by explicitly modifying the inference engine. In this work, we embrace off-policyness and propose a novel off-policy RL algorithm that does not require these modifications: \emph{\underline{O}ptimal \underline{A}dvantage-based \underline{P}olicy Optimization with \underline{L}agged Inference policy} (\algname{}).
We show that \algname{} outperforms GRPO with importance sampling on competition math benchmarks, and can match the performance of a publicly available coding model, DeepCoder, on LiveCodeBench, while using 3x fewer generations during training. We further empirically demonstrate that models trained via \algname{} have improved test time scaling under the Pass@k metric. \algname{} allows for efficient, effective post-training even with lags of more than 400 gradient steps between the training and inference policies, 100x more off-policy than prior approaches.
}

\date{\today}
\correspondence{\email{daniel\textunderscore ritter@seas.harvard.edu}}
\metadata[Code]{\url{https://github.com/danieldritter/OAPL}}

\begin{document}

\maketitle

\section{Introduction}

End-to-end optimization of Large Language Models (LLMs) via Reinforcement Learning (RL) unlocks LLMs' reasoning capabilities. DeepSeek-R1 \citep{Guo2025} is one representative work demonstrating that reasoning capabilities emerge naturally via large-scale RL optimization. Since DeepSeek-R1, the literature has mainly focused on improving the training stability of Group Relative Policy Optimization \citep{shao2024deepseekmathpushinglimitsmathematical} (GRPO) – the RL method that powers the post-training of DeepSeek-R1.

\begin{figure*}[t]
  \vskip 0.2in
  \centering
    \includegraphics[width=0.99\textwidth]{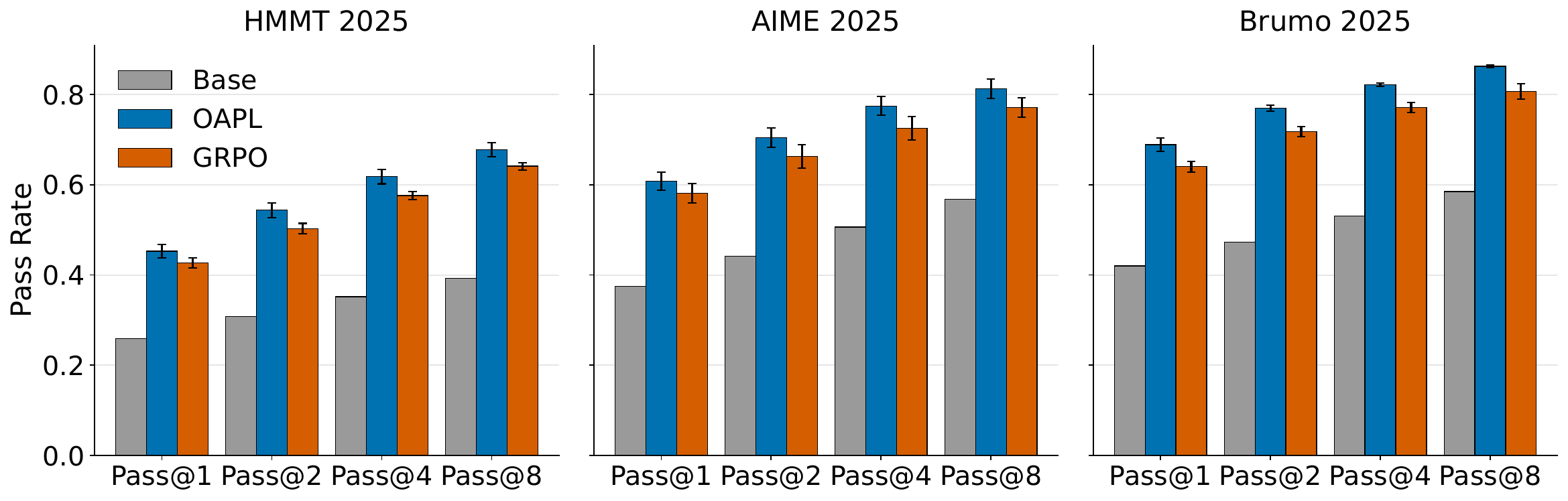}
    \caption{
        \algname{} and GRPO on math reasoning benchmarks.
        Bars show the average of the maximum accuracy across three runs, with error bars indicating
        standard error. We report Pass@1 (computed via averaging over 10 rollouts per prompt), Pass@5, and Pass@10 on (\textbf{Left}) HMMT-25 (Feb \& Nov), (\textbf{Middle}) AIME-25, and (\textbf{Right}) BRUMO-25.}
    \label{fig:math-reasoning-comparison}
\end{figure*}
A central reason training stability is hard to achieve in practice is that modern RL post-training infrastructures are often not truly on-policy.
In particular, the trainer (e.g. a HuggingFace model \citep{wolf2020huggingfacestransformersstateoftheartnatural}) and the inference engine (e.g. a vLLM model \citep{kwon2023efficient}) may produce different log-probabilities for the same sequence, even when both models have the same weights. 
This mismatch can arise because of differences between the trainer and inference engine kernel implementations \citep{yao2025offpolicy,liu-li-2025-rl-collapse,lingteam2025attentionmattersefficienthybrid}, or because of asynchronous training pipelines, where the inference engine may contain an older version of the trainer's weights \citep{fu2025areal}. 

This discrepancy in log-probabilities makes practical policy-gradient training effectively \textbf{off-policy}: the data used to optimize the current policy is not generated by that policy. In contrast, classic policy gradient methods (e.g., REINFORCE \citep{Williams1992}) from which modern policy optimization methods like GRPO and its predecessor PPO \citep{schulman2017proximalpolicyoptimizationalgorithms} are derived, work under the assumption that sampling is \textbf{on-policy}: the data is generated from the current policy to be optimized. 

Most improvements to GRPO thus focus on making it as on-policy as possible despite the gap between the trainer and the inference engine. There are, in general, two families of work tackling this problem: (1) introducing additional importance weights \citep{zheng2025groupsequencepolicyoptimization,yao2025offpolicy,fu2025areal}; (2) reducing the gap between the trainer and inference engine by modifying the inference engine \citep{qi2025defeatingtraininginferencemismatchfp16,nomoretraininginferencemismatch}. While both families demonstrate promising results, we argue that neither is ideal. In the first approach, adding importance weights to the GRPO objective introduces extra variance to the RL loss function.
The second approach makes the inference engine slower and does not fully close the gap between the inference engine and the trainer in asynchronous RL training. 
In this work, we ask the following questions for RL post-training of LLMs: 

\begin{center}
\textbf{Are on-policy algorithms necessary for RL post-training? \\ Can we develop simple and scalable off-policy RL algorithms?}
\end{center}

We find that being on-policy is not necessary for RL post-training, and we propose an easy to implement and effective off-policy post-training algorithm: \emph{\underline{O}ptimal \underline{A}dvantage-based \underline{P}olicy Optimization with \underline{L}agged Inference policy}, abbreviated \algname{}. We treat the mismatch between the trainer and inference engine policies as a KL-regularized RL problem, where the KL term explicitly prevents the training policy from moving too far away from the inference policy. 
Leveraging the closed-form solution of KL-regularized RL, we derive a squared regression objective that trains on rollouts from a lagged inference policy, eliminating the need for on-policy sampling.  \algname{} then uses that objective in an iterative procedure that very infrequently syncs the trainer and inference policy, enabling training that is significantly more off-policy than other approaches. \algname{} fully embraces off-policy training without any importance weighting ratios. Our view that on-policy learning is not necessary for RL post-training is consistent with classical RL results, where on-policy policy gradient methods such as PPO and REINFORCE are often less efficient than off-policy algorithms such as DDPG and SAC \citep{lillicrap2015continuous,haarnoja2018soft} on 
traditional robotics control and video game benchmarks.

Empirically, we find that \algname{} can outperform a GRPO-based baseline on three math competition benchmarks (AIME 25, HMMT 25 Feb and Nov, BRUMO 25) across multiple Pass@k metrics\footnote{To compute these metrics, we sample 10 independent rollouts per prompt. Pass@k is then computed using the unbiased estimator from \citep{chen2021evaluatinglargelanguagemodels}. For our code generation experiments, we sample 20 independent rollouts per prompt and use the same estimator.} (see Figure~\ref{fig:math-reasoning-comparison}). On LiveCodeBench v5, across various Pass@k metrics, our approach can match or outperform DeepCoder \citep{deepcoder2025}, which is trained via GRPO with additional heuristics including clip-high, overlong filtering, etc, while using approximately \textbf{one third the number of generations for training}. Notably, in our code generation experiments, the policy lag (off-policyness) can be as large as 400 gradient updates without the need for any importance sampling. We also observe that \algname{} does not just perform base model distribution sharpening. \algname{} does not cause entropy collapse and stably improves the Pass@k test-time scaling metrics for k ranging from 1 to 256. Overall, we demonstrate that being on-policy is not necessary and embracing off-policy learning can result in stable, effective, and efficient training for reasoning LLMs.

\section{Background}
In modern RL post-training, there are generally two types of policies: the trainer $\pi$ and the inference engine $\pi_{\text{vllm}}$\footnote{We use vLLM as our example inference engine throughout the text}. The trainer $\pi$ is used to compute gradient updates given generated sequences, while the inference engine $\pi_{\text{vllm}}$ is used for fast generation. However, even when $\pi$ and $\pi_{\text{vllm}}$ share the same weights, they can output different log-probabilities given the same sequence of tokens. This inherent difference in the log-probabilities from $\pi$ and $\pi_{\text{vllm}}$ breaks the on-policy assumption of policy gradient-based methods. For example, \citet{liu-li-2025-rl-collapse} measure the KL divergence between the inference engine and trainer, and found that sudden increases in that divergence contributed to training instability and policy collapse in GRPO. The gap between the inference engine and trainer can be further enlarged in an asynchronous RL training framework (e.g., $\pi_{\text{vllm}}$ could be multiple gradient steps behind the trainer $\pi$). \looseness=-1

One common way to handle off-policy rollouts in LLM post-training—and the primary baseline we compare against—is standard importance sampling (IS), applied either at the token level \citep{fu2025areal} or the sequence level \citep{zheng2025groupsequencepolicyoptimization}. Given any prefix $x$ and next token $a$ sampled from $\pi_{\text{vllm}}(\cdot | x)$, IS computes the likelihood ratio $\frac{\pi(a|x)}{\pi_{\text{vllm}}(a|x)}$ and uses it to reweight the GRPO loss function before averaging across a batch of examples. These likelihood ratios aim to correct the mismatch caused by action $a$ being generated from $\pi_{\text{vllm}}$ instead of $\pi$. For example, we can formulate GRPO as an importance weighted loss function:
\begin{equation*}
\label{eq:grpo}
\mathbb{E}_{\{y_i\}_{i=1}^G \sim \pi_{\text{vllm}}(\cdot | x)} \biggl[\frac{1}{G}\sum_{y\in\mathcal{G}} \frac{1}{|y|} \sum_{t=1}^{|y|} \underbrace{ \frac{\pi_{\text{old}}(y_{t}|x,y_{<t})}{\pi_{\text{vllm}}(y_t|x,y_{<t})} }_{\text{IS ratio}} \cdot \min\left\{ r_t A_t, \text{clip}(r_t, 1-\epsilon,1+\epsilon) A_t  \right\}\biggr]
\end{equation*}
where $\pi_{\text{old}}$ is the previous iteration of the trainer, $r_t=\frac{\pi(y_t \mid x, y_{<t})}{\pi_{\text{old}}(y_t \mid x, y_{<t})}$ is the PPO-style likelihood ratio, and $A_t$ is the normalized advantage.
\citet{fu2025areal} introduced this loss function for their asynchronous RL training framework, where the data generation policy $\pi_{\text{vllm}}$ could lag behind the current training policy $\pi$. The additional token-level ratio $\frac{\pi_{\text{old}}(y_t|x,y_{<t})}{\pi_{\text{vllm}}(y_t|x,y_{<t})}$ reweights tokens $y_t$ sampled from $\pi_{\text{vllm}}$ as if they were generated by $\pi_{\text{old}}$.
However, IS can become unreliable when the behavior and target policies differ substantially, motivating a great deal of prior work on variance-reduction techniques \citep{munos2016safeefficientoffpolicyreinforcement,10.5555/3020847.3020905,mahmood2015emphatictemporaldifferencelearning,hallak2015generalizedemphatictemporaldifference,geist2013offpolicylearningeligibilitytraces}. Empirically, prior work has tried many additional heuristics such as clipping  IS ratios, explicitly deleting tokens from the GRPO objective whose IS ratio is too large or too small, or throwing away entire rollouts that are too off-policy. While these heuristics stabilize the GRPO training specifically, they deviate more and more from the principles behind classic policy gradient theory. Since these heuristics are specifically designed for and tested under the GRPO loss, it is unclear how they generalize beyond the very specific GRPO loss function.  In this work, instead of focusing on modifying GRPO's loss, we take a different route, and design a new RL training objective that works naturally with off-policy data.

\section{Method: \algname{}}
We introduce \emph{\underline{O}ptimal \underline{A}dvantage-based \underline{P}olicy Optimization with \underline{L}agged Inference policy} (\algname{}), a principled off-policy objective that remains stable under substantial policy lag. Unlike prior approaches that require inference engine customization or GRPO variants augmented with extra ratios, clipping operators, or deletion of stale tokens/sequences, we embrace the off-policy nature of RL post-training and design a simple, fully off-policy RL algorithm.

\subsection{Off-policy Loss Function}
We first introduce our off-policy policy optimization objective, motivated by the KL-regularized RL formulation. Consider the following objective:
\begin{align}
    \max_{\pi} \mathbb{E}_{x, y\sim \pi(\cdot | x)} r(x,y) - \beta \text{KL}( \pi \vert\vert \pi_{\text{vllm}})
    \label{eq:KL_rej_obj}
\end{align}  
The goal of this objective is to maximize the reward $r$ while at the same time minimizing the KL to the inference policy $\pi_{\text{vllm}}$.\footnote{Note $\pi_{\text{vllm}}$ is not necessarily the reference policy. We use $\pi_{\text{vllm}}$ to denote the current inference policy which could share the weights from the trainer. We emphaize that this is not the usual KL regularization to $\pi_{\text{ref}}$. In fact we do not consider KL regularization to the reference policy in this work. } It is well known that the optimal policy $\pi^\star$ and the optimal value function $V^\star$ of the above KL-regularized RL formulation have the following closed-form expressions:
\begin{align*}
     \pi^\star(y|x) & \propto \pi_{\text{vllm}}(y|x) \exp( r(x,y) / \beta ), \\
      \quad V^\star(x) & = \beta \ln \mathbb{E}_{y\sim \pi_{\text{vllm}}(\cdot | x)} \exp( r(x,y) / \beta ).
\end{align*}
Rearranging terms, we can express the relationship between $\pi^\star$ and $V^\star$ as follows:
\begin{align*}
    \beta \ln \frac{\pi^\star(y|x)}{\pi_{\text{vllm}}(y|x)} = \underbrace{ r(x,y) - V^\star(x)}_{\text{optimal advantage $A^\star$}}, \forall x,y.
\end{align*}

Crucially, the expectation defining $V^\star$ is taken under the sampling policy $\pi_{\text{vllm}}$, not $\pi^\star$.
Thus, given $x$ and a group of $G$ rollouts $\{y_1, \dots, y_G\}$ sampled from $\pi_{\text{vllm}}(\cdot | x)$, \citet{brantley2025accelerating} proposes estimating $V^\star$ by: 
\begin{align}
    \hat V^\star(x) = \beta \ln \frac{1}{G} \sum_{i=1}^G \exp( r(x, y_i) / \beta ).
    \label{eq:vstar}
\end{align}  The estimator $\hat V^\star$ can be accurate under mild assumptions of the sampling distribution $\pi_{\text{vllm}}$. In particular, for a binary reward, if $\pi_{\text{vllm}}$ has a non-zero probability of solving $x$, then $\hat V^\star(x)$ converges to $V^\star(x)$ as $G$ increases \citep{brantley2025accelerating,zhou2025q}. 
The role of $\beta$ here is \emph{smoothing}: when $\beta \to 0$, we have $\hat V^\star(x) = \max_{i} r(x,y_i)$, and when $\beta \to \infty$, $\hat V^\star(x) = \sum_i r(x,y_i) / G$ becomes the average which is an unbiased estimate of the average reward of the current inference policy $\pi_{\text{vllm}}$. 

Given $\hat V^\star$, we can estimate the optimal advantage $A^\star(x,y)$ as $r(x,y) - \hat V^\star(x)$. We adopt the $A^\star$PO objective from \citet{brantley2025accelerating} and define the following policy optimization objective: 
\begin{align}
    \min_{\pi} \sum_{x} \sum_{i=1}^G \left( \beta \ln \frac{ \pi(y_i|x) }{\pi_{\text{vllm}}(y_i|x)} - (r(x,y_i) - \hat V^\star(x)) \right)^2 
    \label{eq:off_policy_apo}
\end{align} 

When $\hat V^\star = V^\star$, Eq.~\ref{eq:off_policy_apo} is minimized by the KL-regularized optimum $\pi^\star$, regardless of the sampling distribution of $y$ (e.g., it holds for rollouts drawn from $\pi_{\text{vllm}}$). While our loss function is motivated by $A^\star$PO, $A^\star$PO was designed to be an on-policy algorithm, i.e., it formulates the above optimization under an on-policy dataset generated from $\pi$. We instead rely on the objective's unique minimizer, and use the off-policy data and log-probabilities from the inference engine directly. 

As motivated by the original $A^\star$PO paper, estimating $\hat{V}^\star$ from groups of rollouts allows us to avoid making extra assumptions, such as $V^\star$ being approximated by a constant \citep{zhu2024starling} or having to use another neural network to model $V^*$, which can be computationally expensive \citep{richemond2024offlineregularisedreinforcementlearning}.

\subsection{\algname: The Off-policy RL Algorithm}
\begin{algorithm}[t]
    \caption{Optimal Advantage-Based Policy Optimization with Lagged Inference Policy (\algname)}
    \label{alg:oapl}
    \begin{algorithmic}
        \Require Policy model $\pi$, Inference engine $\pi_{\text{vllm}}$, Data buffer $\mathcal{D}$, Policy lag $L$
        \State Synchronize $\pi$ and $\pi_{\text{vllm}}$
        \For{$t = 1 \to T$}
            \State Sample a batch $\{x, \{y_i\}_{i=1}^G\}$ from $\pi_{\text{vllm}}$ and store it in $\mathcal{D}$;
            \Comment{\# Data generation (can be async)}
            \State Optimize policy $\pi$ using data from $\mathcal{D}$ via gradient descent on Eq.~\ref{eq:off_policy_apo};
            \Comment{\# Off-policy update (can be async)}
            \If{$t \bmod L = 0$}
                \State Synchronize $\pi_{\text{vllm}}$ with $\pi$ and clear $\mathcal{D}$;
                \Comment{\# Update inference engine}
            \EndIf
        \EndFor
    \end{algorithmic}
\end{algorithm}

We convert Eq.~\ref{eq:off_policy_apo} into a practical post-training pipeline with a lagged inference engine. This yields \emph{\underline{O}ptimal \underline{A}dvantage-Based \underline{P}olicy Optimization with \underline{L}agged Inference Policy} (Algorithm~\ref{alg:oapl}), abbreviated \algname{}. \algname{} begins by synchronizing $\pi$ and $\pi_{\text{vllm}}$ to share the same weights.
The inference engine, using $\pi_{\text{vllm}}$, then begins asynchronously generating data, and adding it to the buffer $\mathcal{D}$. Concurrently, the trainer begins updating the policy $\pi$ by minimizing Eq.~\ref{eq:off_policy_apo}, using data sampled from $\mathcal{D}$. Every $L$ iterations of the trainer, with $L$ being a hyperparameter, the algorithm synchronizes $\pi$ and $\pi_{\text{vllm}}$'s weights. Between synchronizations, the algorithm operates off-policy: $\pi_{\text{vllm}}$ both generates the data and serves as the KL reference in Eq.~\ref{eq:off_policy_apo}. Due to its fully off-policy nature, \algname{} can run completely asynchronously between the two synchronization steps of $\pi$ and $\pi_{\text{vllm}}$.

We clear the buffer $\mathcal{D}$ whenever we synchronize $\pi_{\text{vllm}}$ with $\pi$ to make sure that $\mathcal{D}$ only contains data from a single $\pi_{\text{vllm}}$. This is to ensure that the estimator $\hat V^\star$ and hence the advantage is always computed with data from only one sampling distribution $\pi_{\text{vllm}}$. Because \algname{} does not rely on importance ratios or clipping operations, the resulting update reduces to a simple least-squares regression loss that remains stable even under substantial policy lag.\looseness=-1

\paragraph{Comparison to GRPO} Following PPO's original design, GRPO uses a clipping operator on $\frac{\pi(y|x)}{\pi_{\text{old}}(y|x)}$ to prevent $\pi$ from deviating too far from $\pi_{\text{old}}$ -- the trainer from the previous iteration. This is motivated by conservative policy iteration \citep{kakade2002approximately}. However, clipping is not always effective at preventing $\pi$ from deviating from $\pi_{\text{old}}$. When beginning with $\pi = \pi_{\text{old}}$, the computation of the first gradient update using the GRPO loss does not cause any clipping. Thus if the first gradient is large, one step of gradient descent could already take $\pi$ far away from $\pi_{\text{old}}$, and the clipping operator cannot pull $\pi$ back to $\pi_{\text{old}}$. This is a known problem of PPO/GRPO's loss functions \citep{hsu2020revisiting}. 
In contrast, \algname{} incorporates KL regularization to $\pi_{\text{vllm}}$ into the optimization objective,   completely abandons the concept of $\pi_{\text{old}}$, and directly uses the log-probabilities from the sampling distribution $\pi_{\text{vllm}}$. Thus, in each iteration, \algname{} directly encourages the trainer $\pi$ to stay close to $\pi_{\text{vllm}}$ while optimizing the reward.  As we will show in the experiments, this design, together with the infrequent updates of $\pi_{\text{vllm}}$, can keep the entropy of the policy from collapsing during training, leading to better test time scaling than GRPO.

\paragraph{Comparison to $A^\star$PO} $A^\star$PO was originally designed as an on-policy RL algorithm, and it estimates $V^\star$ defined under a fixed reference policy, $\pi_{\text{ref}}$, using $\ln \frac{\pi(y|x)}{\pi_{\text{ref}}(y|x)}$ inside the loss function. It never updates $\pi_{\text{ref}}$ during training. In contrast, \algname{} runs in an off-policy manner, periodically updates the inference engine $\pi_{\text{vllm}}$, and always uses the log-probabilities from $\pi_{\text{vllm}}$ directly in the loss function. 

\section{Related Work}
\paragraph{Off-policy RL Post-Training}
Approaches for dealing with off-policy sampling in RL post-training can broadly be divided into those that avoid importance sampling, and those that apply importance sampling or a related variation of it.

Examples of methods that avoid importance sampling include \citet{melo2025stabilizingpolicygradientssampleefficient}, who estimate Fisher information for token masking, or \citet{arnal2025asymmetricreinforceoffpolicyreinforcement}, who bias their objective function for performance improvement guarantees. \algname{} similarly avoids the added variance of importance sampling, but does not require additional estimation procedures while remaining unbiased. The most closely related works to ours in this category use squared regression losses for either on- or off-policy training, e.g. REBEL \citep{gao2024rebel}, REFUEL \citep{gao2024regressing}, AGRO \citep{tang2025rlfinetuningllmsonoffpolicy} or Kimi K2 \citep{kimiteam2025kimik2openagentic}. However, these approaches do not  estimate $V^*$ as OAPL does, replacing it instead with group-relative baselines for variance reduction, similar to the RLOO estimator \citep{Kool2019Buy4R}.

Approaches that rely on importance sampling, or just the importance ratio $\frac{\pi(y|x)}{\pi_{\text{vllm}}(y|x)}$, vary in exactly how they apply it. For instance, DeepSeek-v3.2 \citep{liu2025deepseek} deletes rollouts whose likelihood is small under $\pi$, and \citet{IcePop2025} and \citet{zheng2025prosperitycollapsefaroffpolicy} delete tokens whose token-level ratio is too large or too small. \citet{roux2025taperedoffpolicyreinforcestable} and \citet{su2026klearreasoneradvancingreasoningcapability} construct objective functions to bound the gradients of tokens with large importance ratios. By avoiding importance sampling, \algname{} avoids having to delete samples or tokens that could be useful for learning, and does not incur bias or additional tuning costs by adding clipping to ratios or gradients.

\paragraph{Off-Policy RL in Asynchronous Settings} Work on asynchronous and large-scale RL training has also dealt with off-policy sampling. Recent work on scaling up RL from human feedback systems \citep{noukhovitch2025asynchronousrlhffasterefficient,khatri2025artscalingreinforcementlearning}, for instance, has used truncated importance sampling for this issue. Outside of the language modeling context, methods for scaling up policy gradient algorithms and leveraging off-policy data have also used some form of (usually truncated) importance sampling \citep{espeholt2018impalascalabledistributeddeeprl,munos2016safeefficientoffpolicyreinforcement,wang2017sampleefficientactorcriticexperience,10.5555/3020847.3020905}, or have constrained their data generation to avoid collecting data that is too off-policy \citep{openai2019dota2largescale}. Other approaches avoid importance sampling entirely by learning a Q-function \citep{haarnoja2018soft,Mnih2015,mnih2016asynchronousmethodsdeepreinforcement}. \algname{} similarly requires no importance sampling, and can actually be understood as a value learning approach which uses $\ln \frac{\pi}{\pi_{\text{vllm}}}$ as a function approximator to estimate the optimal advantage $A^\star$ directly.

\section{Experimental Setup}
We evaluate \algname{} on competition mathematical problem solving and code generation, focusing on stability during asynchronous training and performance measured by Pass@k.
For both the competition math and code generation settings, as in \citet{brantley2025accelerating}, we use two separate betas-$\beta_1,\beta_2$, in Equations \ref{eq:vstar} and \ref{eq:off_policy_apo}, respectively, rather than a single $\beta$. This allows for additional freedom in choosing hyperparameters. Additional details about the training setup and hyperparameters for the experiments can be found in Appendix \ref{experimental_details_appendix}.

\subsection{Math Experimental Setup}
For our math experiments, we use Deepscaler \citep{deepscaler2025} as our training dataset and AIME 25, HMMT 25 (Feb and Nov), and BRUMO 25 as our evaluation sets. We compare \algname{} to GRPO with additional importance sampling that accounts for the log-probability difference between the inference engine and the trainer \citep{yao2025offpolicy}. For both approaches, we implement asynchronous optimization, which means that the inference engine can generate data while we optimize the trainer. For \algname{}, we set $L = 50$, meaning that we synchronize the inference engine and the trainer every 50 iterations. For GRPO, we use off-by-one asynchronous training. Namely, the training data used by the trainer can come from an inference policy that is at most 1 iteration older than the trainer itself. We use Qwen3-4B-Thinking-2507 as our base model, with a maximum generation length of 16384 tokens for both methods.

\subsection{Code Generation Experimental Setup}
\label{coding-experimental-setup}
For the code generation experiments, we use a highly off-policy two-stage training process to replicate the performance of DeepCoder \citep{deepcoder2025}, a publicly available coding model trained via GRPO with several additional heuristics. Beginning with the base model, DeepSeek-R1-Distill-Qwen-14B, we generate an offline dataset of 8 responses for every prompt in DeepCoder’s training dataset. To focus training on feasible problems, we additionally filter out all prompts where the model generated no correct responses. We then train the base model on this dataset with OAPL for 1 epoch without synchronizing the trainer and inference engines. Using the resulting model, we generate a new offline dataset from a random subset of 4000 prompts (due to resource constraints), and continue training for an additional four epochs on this dataset. This is equivalent to running OAPL with $L$ set to 1 epoch (approximately 400 gradient updates) and total iterations $T = 2$. The maximum generation length for both rounds of training is 32K tokens. For evaluation, we follow DeepCoder's LiveCodeBench \citep{jain2024livecodebench} setup, using the same subset of 279 LiveCodeBench problems, and evaluating with a maximum generation length of 64K. We evaluate all four checkpoints from each epoch of the secound round of training for \algname{}, and report results for the best performing checkpoint.\footnote{The offline datasets and checkpoints from the first and second rounds of training are available \href{https://huggingface.co/collections/danieldritter/oapl}{here}, and the training code for the code generation experiments is available \href{https://github.com/danieldritter/OAPL}{here}. Checkpoints for our math experiments will be available soon.}

\section{Experimental Results}

\begin{figure*}[t]
    \centering
    \begin{subfigure}[t]{0.32\linewidth}
        \centering
        \includegraphics[width=\linewidth]{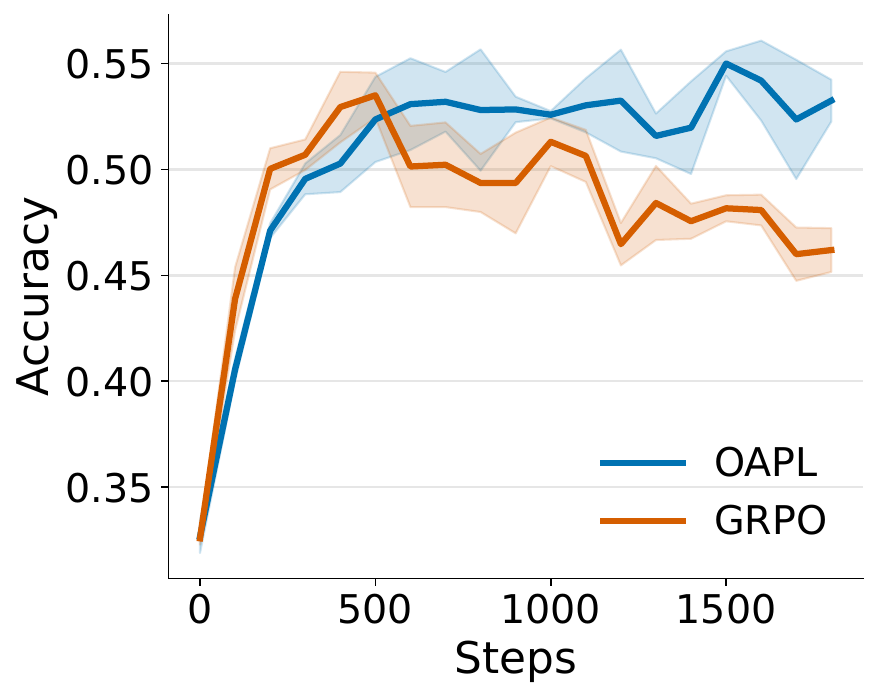}
    \end{subfigure}
    \hfill
    \begin{subfigure}[t]{0.32\linewidth}
        \centering
        \includegraphics[width=\linewidth]{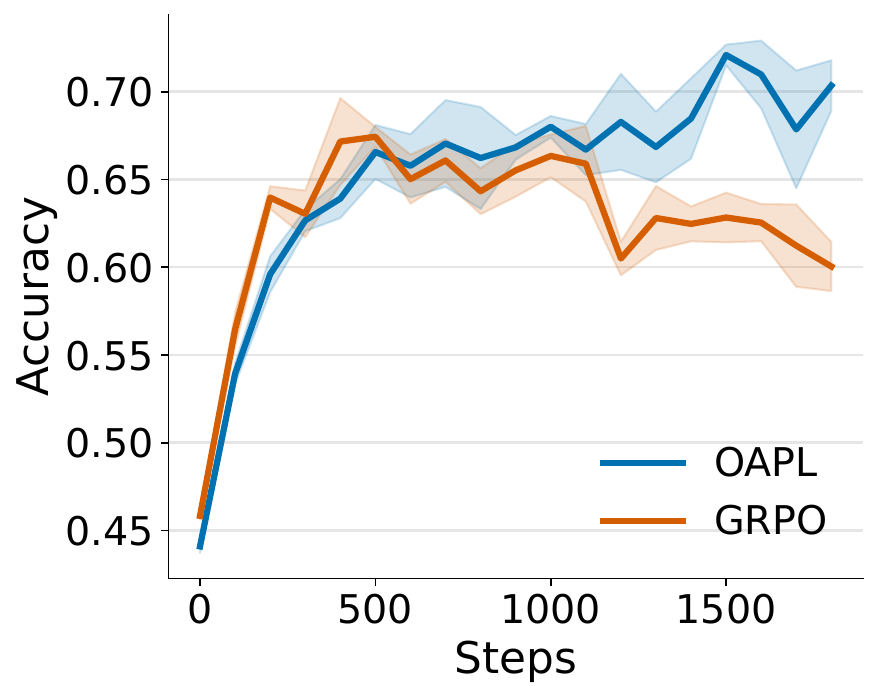}
    \end{subfigure}
        \hfill
    \begin{subfigure}[t]{0.32\linewidth}
        \centering
        \includegraphics[width=\linewidth]{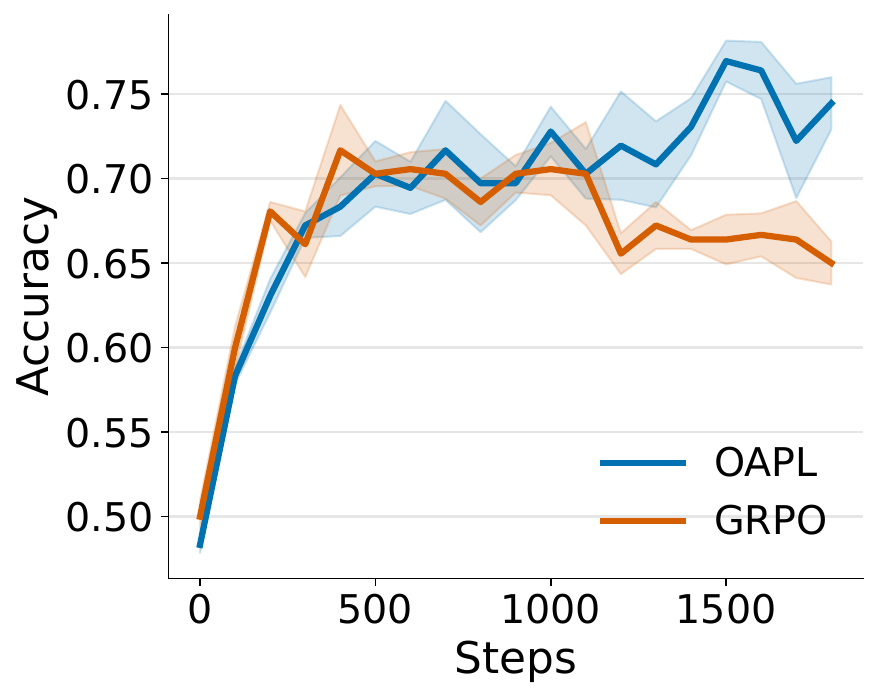}
    \end{subfigure}
    \caption{Training curves on competition math. Curves show mean accuracy across three benchmarks (AIME25, HMMT25, BRUMO25) and shaded regions denote standard error. (\textbf{Left}) Pass@1, (\textbf{Middle}) Pass@5, and (\textbf{Right}) Pass@10. \algname{} converges to higher accuracy and remains more stable than GRPO over training.}
    \label{fig:math_passatk}
\end{figure*}

We evaluate \algname{} along three axes: final accuracy on standard reasoning benchmarks, training dynamics and stability under asynchronous rollouts, and test-time scaling as measured by Pass@k. We first study competition math, where we can track learning curves and entropy over training, and then turn to code generation, where we evaluate robustness under \emph{extreme policy lag} and compare against the GRPO-trained DeepCoder model.

\subsection{Results on Competition Math}

\paragraph{Performance on benchmarks} Figure~\ref{fig:math-reasoning-comparison} demonstrates that \algname{} outperforms the GRPO baseline on all three benchmarks across Pass@k for various $k$. Figure ~\ref{fig:math_passatk} additionally shows the performance across training, averaged over all three benchmarks. Overall, we see that \algname{} learns more stably than and outperforms GRPO. We also observe that, for both GRPO and \algname{}, training on the Pass@1 reward (i.e., the outcome reward alone) improves the Pass@k for $k>1$. Including the code generation experiments that we will show, we in general do not observe the phenomenon that RL does not improve Pass@k for $k>1$.  
\paragraph{Entropy behavior} 

\begin{figure*}[t]
    \centering
    \begin{subfigure}[t]{0.48\textwidth}
        \centering
        \includegraphics[width=\linewidth]{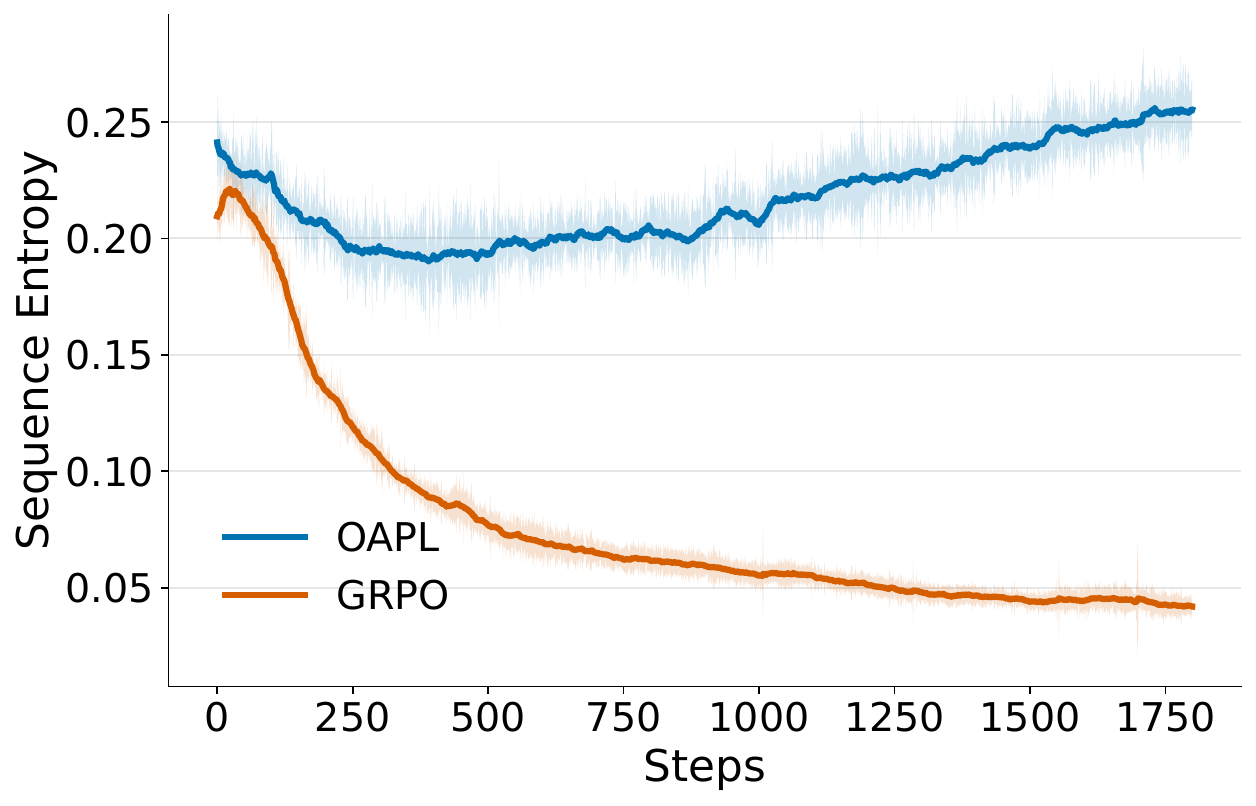}
        %\caption{\kiante{Entropy}}
        \label{fig:entropy}
    \end{subfigure}
    \hfill
    \begin{subfigure}[t]{0.48\textwidth}
        \centering
        \includegraphics[width=\linewidth]{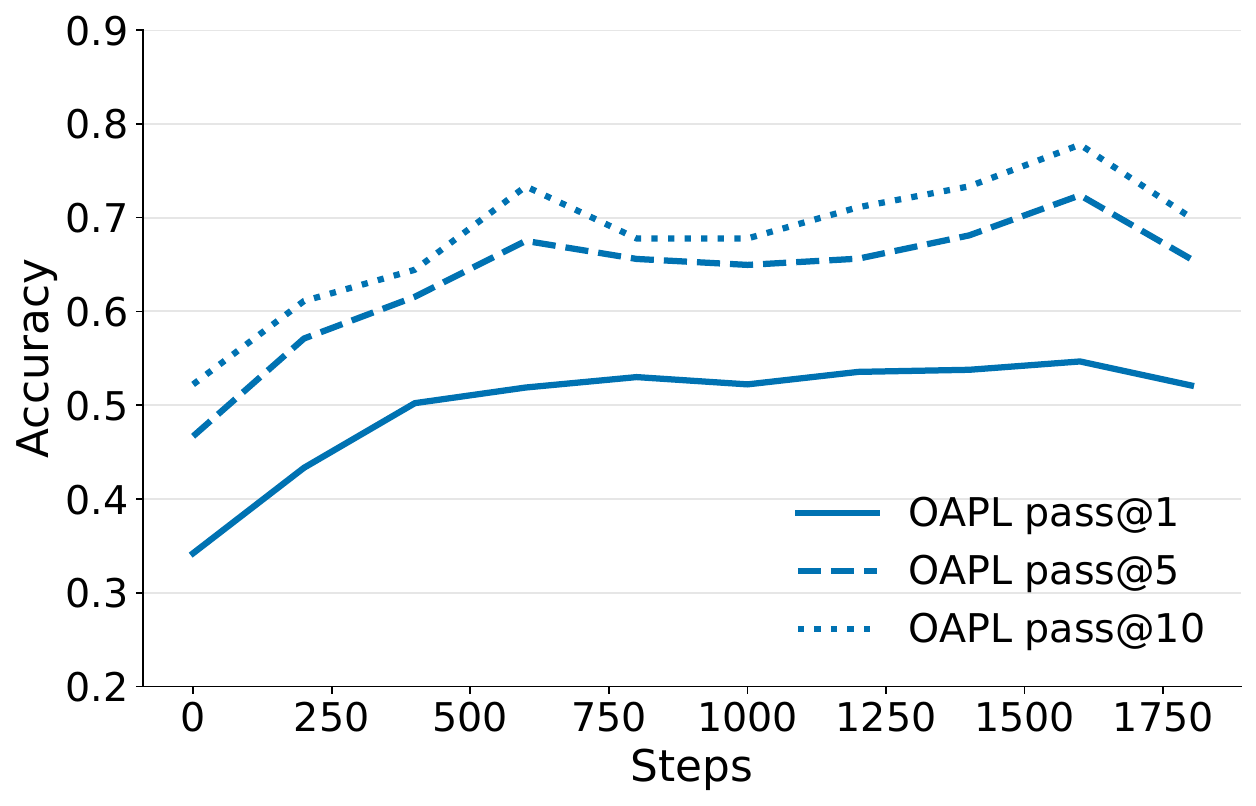}
        %\caption{\kiante{Accuracy ($K=100$)}}
        \label{fig:k_100}
    \end{subfigure}

    \caption{
        Training dynamics and robustness to policy lag in competitive math. (\textbf{Left}) The training entropy for both \algname{} and GRPO (mean across three runs; shaded region is standard error). (\textbf{Right}) Accuracy over training for \algname{} with a larger synchronization interval ($K=100$), averaged over AIME, HMMT, and BRUMO; dashed/dotted lines show Pass@1/5/10 computed from 10 rollouts per prompt. \algname{} remains stable even under substantially lagged rollouts.
    }
    \label{fig:training_dynamics}
\end{figure*}

Figure~\ref{fig:training_dynamics} (left) shows the change in sequence entropy during training. We observe that \algname{}'s entropy does not collapse while GRPO's does. The increased entropy we observe from training with OAPL contributes to the improved performance of \algname{} over GRPO on the Pass@5 and Pass@10 metrics in Figure.~\ref{fig:math_passatk}. We believe this behavior is due to the infrequent synchronization between the inference engine and trainer and to the explicit KL regularization of the trainer against the inference engine.
Note that in our experiments neither GRPO nor \algname{} include a fixed KL regularization to $\pi_{\text{ref}}$ -- the original pre-trained policy (e.g., Qwen3-4B-thinking in this case). This is because the goal of both \algname{} and the GRPO baseline is just to find the policy that optimizes the reward.  
\paragraph{Scaling $k$ in Pass@k} 

\iffalse
\begin{figure*}[t!]
\centering
  \centering
  \begin{subfigure}[t]{\linewidth}
    \centering
    \includegraphics[width=\linewidth]{figures/passkscale_average.pdf}
    \caption{Average}
    \label{fig:avg_pass_k_scal}
  \end{subfigure}
\hfill
  \centering
  \begin{subfigure}[t]{0.49\linewidth}
    \centering
    \includegraphics[width=\linewidth]{figures/passkscale_AIME25.pdf}
    \caption{AIME 25}
    \label{fig:aime_pass_k_scal}
  \end{subfigure}
  \hfill
  \begin{subfigure}[t]{0.49\linewidth}
    \centering
    \includegraphics[width=\linewidth]{figures/passkscale_HMMT25-nov.pdf}
    \caption{HMMT 25 Nov}
    \label{fig:hmmtnov_pass_k_scal}
  \end{subfigure}

  \vspace{0.6em}

  \begin{subfigure}[t]{0.49\linewidth}
    \centering
    \includegraphics[width=\linewidth]{figures/passkscale_HMMT25-feb.pdf}
    \caption{HMMT 25 Feb}
    \label{fig:hmmtfeb_pass_k_scal}
  \end{subfigure}
  \hfill
  \begin{subfigure}[t]{0.49\linewidth}
    \centering
    \includegraphics[width=\linewidth]{figures/passkscale_BRUMO25.pdf}
    \caption{BRUMO 25}
    \label{fig:brumo_pass_k_scal}
  \end{subfigure}
\caption{
        Scaling behaviors of \algname{} and GRPO for Pass@k. We  observe RL training increases Pass@k for all $k$ ranging from $1$ to $256$. \algname{} improves scaling relative to GRPO and the base model. (\textbf{Left}) Average across all benchmarks; remaining panels show per-benchmark results (AIME25, HMMT25 Nov, HMMT25 Feb, BRUMO25).
}
\label{fig:pass_k_scale}
\end{figure*}
\fi

\begin{figure*}[t!]
\centering

\begin{subfigure}[t]{0.32\textwidth}
    \centering
    \includegraphics[width=\linewidth]{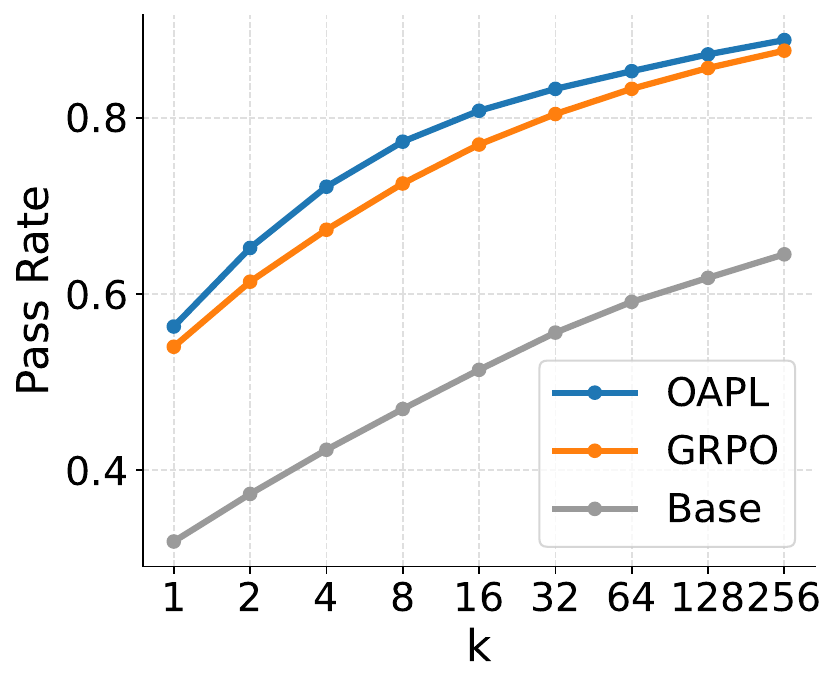}
    \caption{Average}
    \label{fig:avg_pass_k_scal}
\end{subfigure}
\begin{subfigure}[t]{0.32\textwidth}
    \centering
    \includegraphics[width=\linewidth]{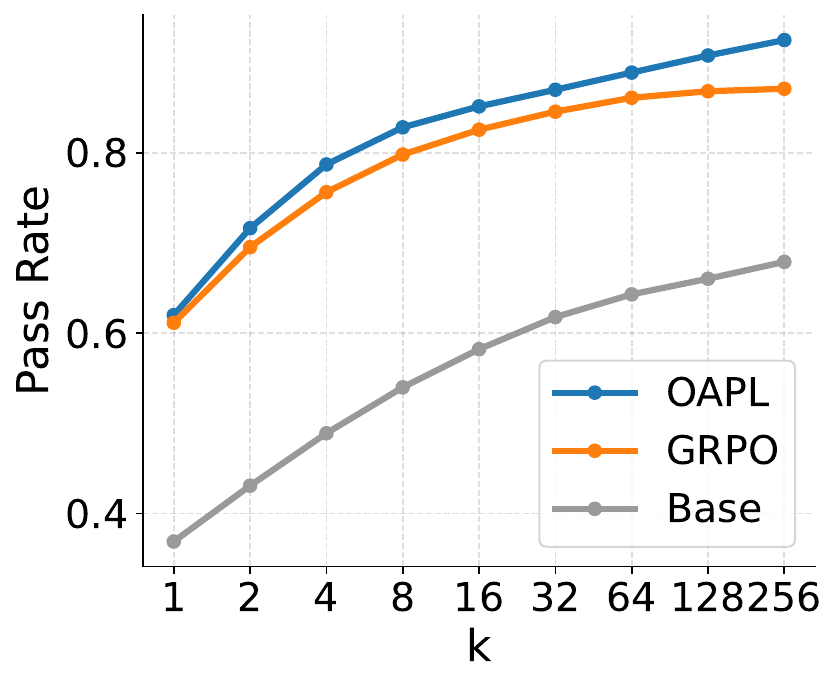}
    \caption{AIME 25}
    \label{fig:aime_pass_k_scal}
\end{subfigure}
\begin{subfigure}[t]{0.32\textwidth}
    \centering
    \includegraphics[width=\linewidth]{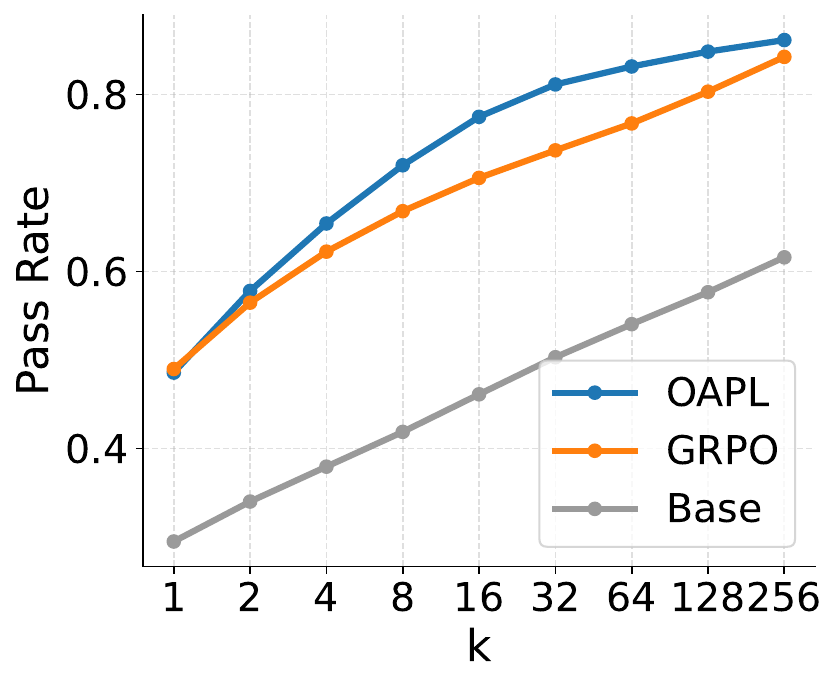}
    \caption{HMMT 25 Nov}
    \label{fig:hmmtnov_pass_k_scal}
\end{subfigure}

\begin{subfigure}[t]{0.32\textwidth}
    \centering
    \includegraphics[width=\linewidth]{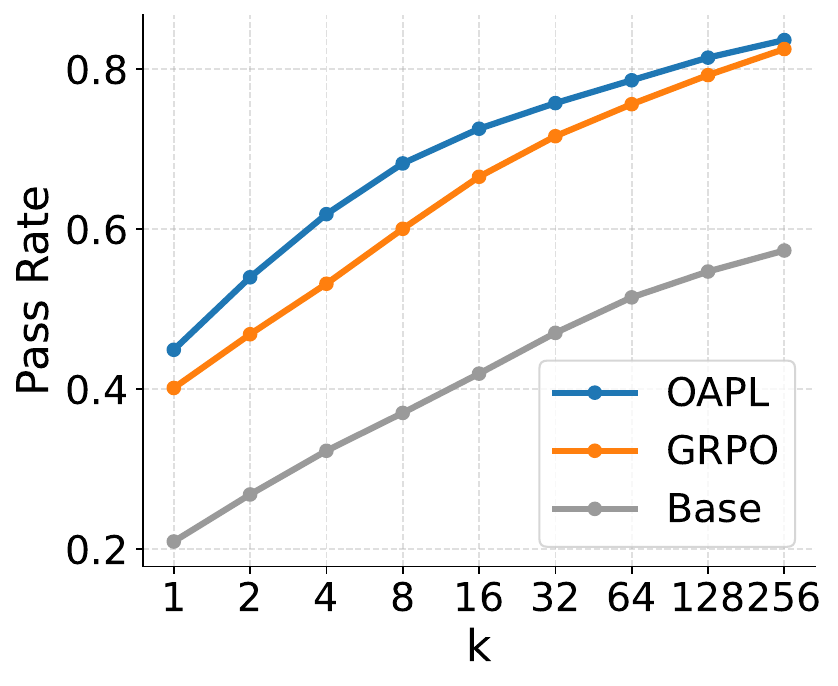}
    \caption{HMMT 25 Feb}
    \label{fig:hmmtfeb_pass_k_scal}
\end{subfigure}
\begin{subfigure}[t]{0.32\textwidth}
    \centering
    \includegraphics[width=\linewidth]{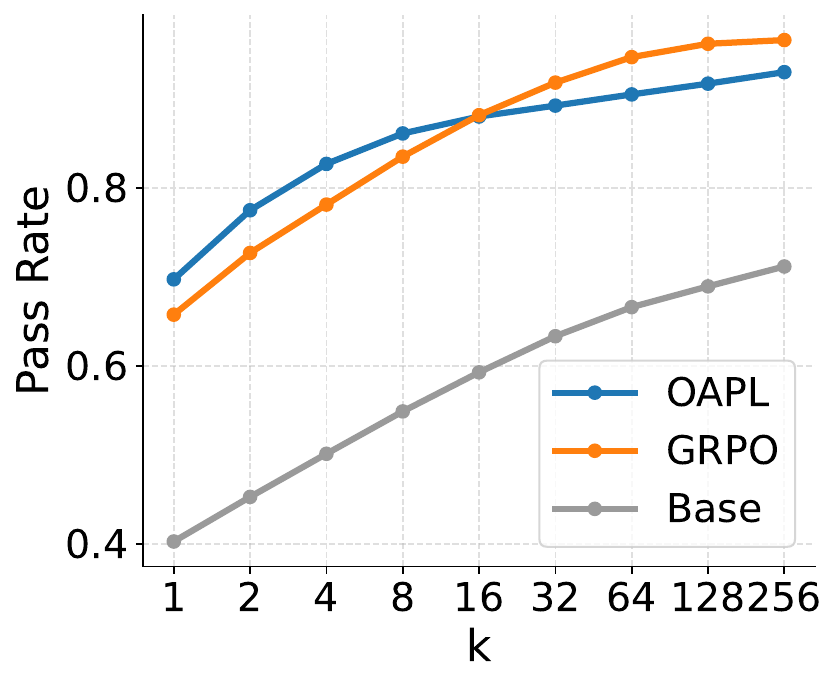}
    \caption{BRUMO 25}
    \label{fig:brumo_pass_k_scal}
\end{subfigure}

\caption{
Scaling behaviors of \algname{} and GRPO for Pass@k. We observe RL training increases Pass@k for all $k$ ranging from $1$ to $256$. \algname{} improves scaling relative to GRPO and the base model. (\textbf{Left}) Average across all benchmarks; remaining panels show per-benchmark results (AIME25, HMMT25 Nov, HMMT25 Feb, BRUMO25).
}
\label{fig:pass_k_scale}
\end{figure*}

Does higher entropy in \algname{} lead to better scaling behavior under Pass@k? We select the best checkpoint for each method (based on the average Pass@1 over the three benchmarks) and evaluate Pass@k as $k$ increases.
Figure~\ref{fig:pass_k_scale} demonstrates that $\algname{}$ scales better than GRPO on average (left), and on every benchmark except BRUMO where both methods already achieve accuracy above $90$ at $k=64$. In particular, we observe a large gap between \algname{} and GRPO for HMMT Nov 2025. Interestingly, \textbf{we observe that RL training (\algname{} and GRPO) improves Pass@k across a wide range of $k$ compared to the base model (e.g., on HMMT 25 Nov, the gap between \algname{} and base actually increases as $k$ increases)}. 
This is in sharp contrast to much prior work (e.g. \citet{yue2025doesreinforcementlearningreally}) arguing that RL only sharpens the base model distribution, in the sense that it does not improve Pass@k for large k. 
\paragraph{Training stability with large policy lags} 
Can \algname{} still learn stably when the inference engine policy lags significantly behind the trainer?  We further evaluate \algname{} with $L=100$, i.e., we only synchronize $\pi_{\text{vllm}}$ and $\pi$ every 100 iterations. As shown in Figure.~\ref{fig:training_dynamics} (Right), \algname{} continues to learn stably, which demonstrates the robustness of \algname{} to different levels of off-policyness in the training data. 
\subsection{Results on Code Generation}

\begin{figure*}[t]
    \centering
    \begin{subfigure}[t]{0.49\textwidth}
        \centering
        \includegraphics[width=1.\linewidth]{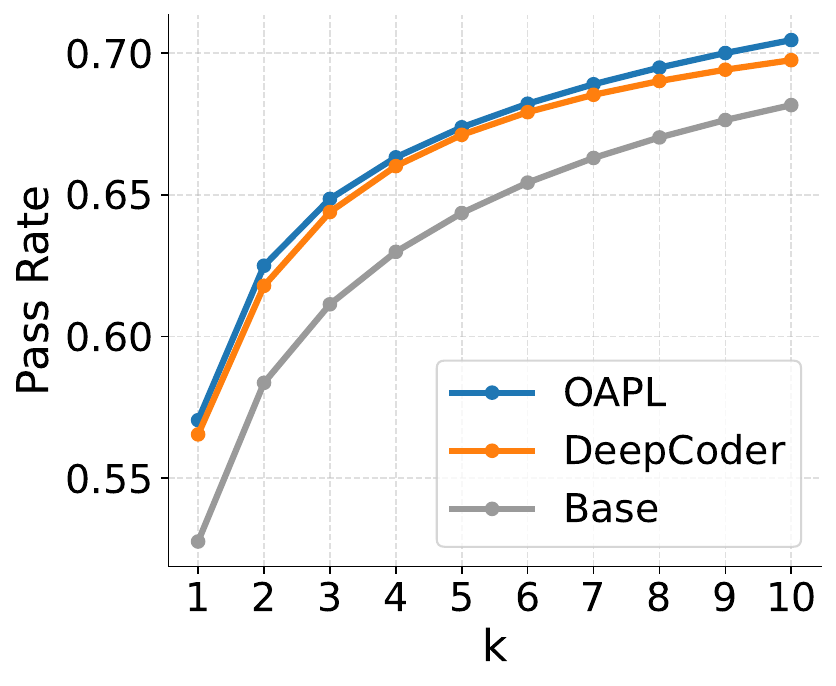}
        %\caption{\kiante{Pass@k scaling}}
    \end{subfigure}
    \hfill
    \begin{subfigure}[t]{0.49\textwidth}
        \centering
        \includegraphics[width=1.\linewidth]{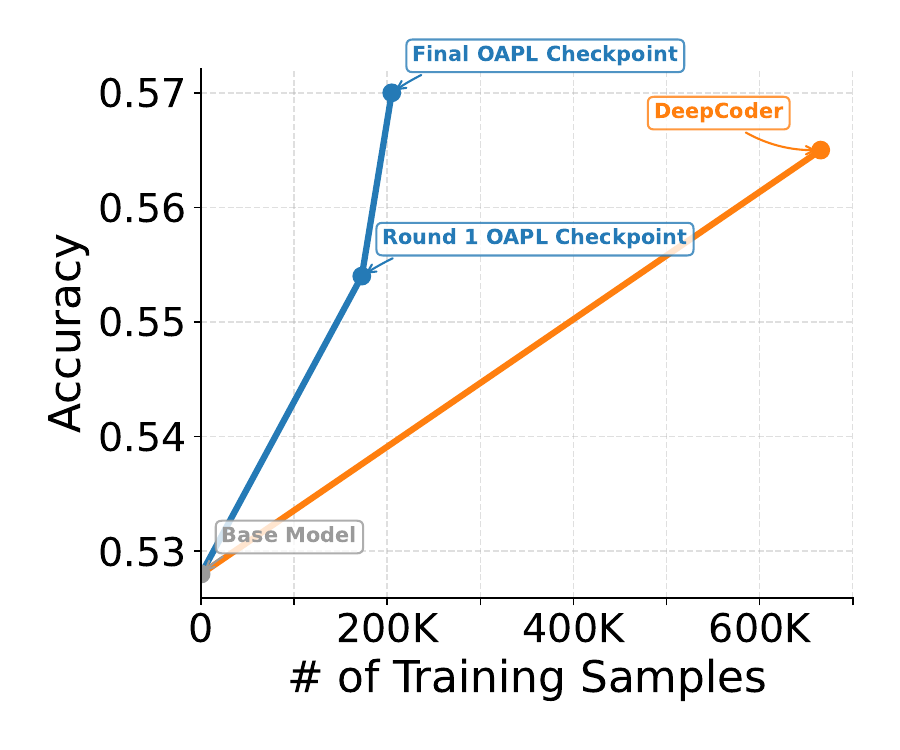}
        %\caption{\kiante{Sample efficiency}}
    \end{subfigure}

    \caption{
    Code generation results on LiveCodeBench. (\textbf{Left}) Pass@k scaling for \algname{}, DeepCoder, and the shared base model. (\textbf{Right}) Sample efficiency: Pass@1 Accuracy versus the number of training generations, highlighting that \algname{} matches DeepCoder while using substantially fewer samples. All metrics are computed from 20 rollouts per prompt using the same evaluation protocol as DeepCoder.}
    \label{fig:deepcoder_results}
\end{figure*}
We evaluate whether \algname{} remains effective under extreme off-policyness using the two-stage offline rollout procedure described in Section \ref{coding-experimental-setup}, and compare against DeepCoder \citep{deepcoder2025} on LiveCodeBench.
\paragraph{Pass@k performance.}
Figure~\ref{fig:deepcoder_results} (Left) shows the Pass@k performance on LiveCodeBench for DeepCoder, our OAPL-trained replication model, and the base model used for both (Deepseek-R1-Distill-Qwen-14B) \footnote{Note that our DeepCoder accuracy is lower than the originally reported 60.6\% accuracy. We were unable to replicate the 60.6\% result despite our best efforts to match their sampling parameters and software environment. See \href{https://github.com/rllm-org/rllm/issues/113}{here} for an open issue where others have had difficulty reproducing the result as well. We also use 20 samples per prompt when computing all Pass@k to reduce the potential randomness in evaluation. }. Pass@k increases with $k$ for all models. Across the entire range of $k$, the \algname{}-trained model matches or slightly outperforms DeepCoder. Comparing to the scaling curve from the base model, we again see that RL training (both \algname{} and the GRPO variant used for DeepCoder)  improves pass@k for large $k$.

\paragraph{Sample efficiency.}
Training with OAPL is also significantly more sample efficient than the original DeepCoder training pipeline. Figure~\ref{fig:deepcoder_results} (Right) shows OAPL and DeepCoder's Pass@1 performance as a function of total training samples. DeepCoder used approximately 650K samples during training\footnote{Based on the training runs released \href{https://wandb.ai/mluo/deepcoder}{here}, DeepCoder was trained for 650 steps, with 1024 samples generated per step.}. In contrast, training with OAPL required only $\sim$200K samples. This represents an approximately 3x reduction in the number of samples required, while achieving equal or better performance. This comparison does slightly inflate the actual total computational cost of DeepCoder, as the first part of their training (160 steps) is limited to 16K length generations, and switches to 32K later. But even if we count the 16K generations as `half' a sample for a fairer accounting, the total is approximately 580K samples, and OAPL still provides significant sample efficiency gains.

\section{Conclusion and Future Work}
Our work demonstrates that a simple off-policy RL method can be more effective than GRPO, an on-policy RL method for LLM post-training. Off-policy methods enable fully asynchronous training and allow algorithms to reuse previously sampled data, often yielding superior computational and sample efficiency. Our experimental results show that we can achieve equivalent or better performance to GRPO on competition math and code generation tasks, in addition to significant increases in sample efficiency from training off-policy. We are excited to continue exploring off-policy training, including training value functions in an off-policy manner for better credit assignment, and leveraging additional offline data (e.g., human data) for more efficient RL. 

\section*{Acknowledgement}
KB acknowledges the Chan Zuckerberg Initiative Foundation for establishing the Kempner Institute for the Study of Natural and Artificial Intelligence. DR acknowledges the support of Schmidt Sciences Humanities and AI Virutal Institute.

\newpage
\bibliography{reference}
\newpage 
\appendix
\section{Experimental Details}
\label{experimental_details_appendix}
We include the detailed hyperparameters of the algorithms here. 
\subsection{Math Training Hyperparameters}
\begin{table}[h]
\centering
\label{tab:optimizer-hparams}
\begin{tabular}{ll}
\toprule
\textbf{Parameter} & \textbf{Value} \\
\midrule
Optimizer              & AdamW \\
Learning rate          & $1\times10^{-6}$ \\
$\beta_1$              & 0.9 \\
$\beta_2$              & 0.95 \\
Weight decay           & $1\times10^{-2}$ \\
Gradient clipping type & Norm \\
Clipping threshold     & $1\times10^{-3}$ \\
\bottomrule
\end{tabular}
\caption{Optimizer hyperparameters used for both \algname{} and GRPO for the math task. Note that $\beta_1$ and $\beta_2$ here are for AdamW, not \algname.}
\label{table:optimizer}
\end{table}

Table~\ref{table:optimizer} shows the hyperparameters we use for the optimizer for both methods. We did not tune the optimizer for the math task. Table~\ref{table:hp_1} shows the method-specific hyperparameters, and Table~\ref {table:hp_2} shows the shared hyperparameters of both approaches. For \algname{}, we did hyperparameter search over $\beta_1 = \{1,5\}$ and $\beta_2 = \{1e-2,1e-3\}$. We observe that $\{\beta_1 = 1, \beta_2 = 1e-3\}$ gives the best overall performance (on average), and report performance with those values.

\begin{table}[h]
\centering

\label{tab:method-hparams}
\begin{tabular}{lll}
\toprule
\textbf{Parameter} & \textbf{OAPL} & \textbf{GRPO} \\
\midrule
$\beta_1$              & 1            & -- \\
$\beta_2$              & $1\times10^{-3}$ & -- \\
$L$                    & 50           & -- \\
Clip ratio             & --           & 0.2 \\
Length normalization   & --           & True \\
Max async iterations    & --      & 1 \\
\bottomrule
\end{tabular}
\caption{Method-specific hyperparameters for \algname{} and GRPO on math task.}
\label{table:hp_1}
\end{table}

\begin{table}[h]
\centering
\label{tab:shared-hparams}
\begin{tabular}{lll}
\toprule
\textbf{Category} & \textbf{Parameter} & \textbf{Value} \\
\midrule
Training
& Generations per prompt (G)      & 8 \\
& Batches per update          & 2 \\
& Global train batch size     & 128 \\
\midrule
Evaluation
& Temperature                 & 0.6 \\
& Top-$p$                     & 0.95 \\
& Max tokens                  & 16384 \\
\midrule
Generation
& Temperature                 & 1.0 \\
& Top-$p$                     & 1.0 \\
& Max tokens                  & 16384 \\
\bottomrule
\end{tabular}
\caption{Shared hyperparameters for \algname{} and GRPO.}
\label{table:hp_2}
\end{table}

\subsection{Code Generation Training Hyperparameters}
\begin{table}[H]
\centering

\begin{tabular}{ll}
\toprule
\textbf{Parameter} & \textbf{Value} \\
\midrule
Optimizer & AdamW \\
$\beta_1$ & 0.9 \\
$\beta_2$ & 0.999 \\
Weight decay & $1 \times 10^-2$ \\
Gradient clipping type & Norm \\
Gradient clipping threshold & 1.0 \\

\bottomrule
\end{tabular}
\caption{Optimizer hyperparameters for code generation experiments}
\label{table:coding-optimizer-hparams}

\end{table}

\begin{table}[H]
\centering

\begin{tabular}{ll}
\toprule
\textbf{Parameter} & \textbf{Value} \\
\midrule
$\beta_1$              & 1  \\
$\beta_2$              & $1\times10^{-3}$ \\
$L$                    & 418 \\
\bottomrule
\end{tabular}
\caption{OAPL Hyperparameters for code generation experiments}
\label{table:coding-oapl-hparams}

\end{table}

\begin{table}[h]
\centering
\label{tab:shared-hparams-code}
\begin{tabular}{lll}
\toprule
\textbf{Category} & \textbf{Parameter} & \textbf{Value} \\
\midrule
Training
& Generations per prompt (G)      & 8 \\
& Batches per update          & 1 \\
& Global train batch size     & 256 \\
\midrule
Evaluation
& Temperature                 & 0.6 \\
& Top-$p$                     & 0.95 \\
& Max tokens                  & 65536 \\
\midrule
Generation
& Temperature                 & 1.0 \\
& Top-$p$                     & 1.0 \\
& Max tokens                  & 32000 \\
\bottomrule
\end{tabular}
\caption{Training and Eval hyperparameters for code generation experiments}
\label{table:coding_training_hparams}
\end{table}
Tables \ref{table:coding-optimizer-hparams}, \ref{table:coding-oapl-hparams}, and \ref{table:coding_training_hparams} show the optimizer, \algname{}-specific, and training hyperparameters, respectively, for our code generation experiments. We did not sweep to choose hyperparameters, due to the computational cost of runs, and chose $\beta_1,\beta_2$ for \algname{} based on defaults found to be effective in other experiments.

\end{document}

%% file: header.tex
\usepackage{url}
\usepackage{wrapfig}
\usepackage{makecell}
\usepackage{enumitem}
\usepackage{tikz}
\usepackage{mathtools}
\usepackage{listings}
\usepackage{algorithm}
\usepackage{algorithmicx}
\usepackage[noend]{algpseudocode}

\definecolor{dbxfg}{HTML}{1C2B33}
\definecolor{dbxbg}{HTML}{F1F4F7}
\definecolor{accentblue}{HTML}{376DCC}
\definecolor{accentorange}{HTML}{EE5A24}

\lstdefinelanguage{json}{
  basicstyle=\ttfamily\footnotesize,
  numbers=left,
  stepnumber=1,
  numberstyle=\tiny,
  breaklines=true,
}

\lstdefinestyle{mdsrc}{
  basicstyle=\ttfamily\scriptsize,
  breaklines=true,
  numbers=left,
  numberstyle=\tiny
}

\tcbset{
  templatebox/.style={
    colback=dbxbg,
    colframe=black,
    enhanced,
    boxrule=0.6pt,
    arc=2pt,
    left=6pt,
    right=6pt,
    top=6pt,
    bottom=6pt
  }
}
\newtcolorbox{TemplateBox}[1][]{templatebox,#1}

\algrenewcommand\algorithmicrequire{\textbf{Input:}}
\algrenewcommand\algorithmicensure{\textbf{Output:}}

%% file: notation.tex
\newcommand{\algname}{OAPL}